\documentclass[11pt,a4paper,logo]{googledeepmind}

\ifdefined\pdfinfoomitdate
\fi
\ifdefined\pdftrailerid
  \pdftrailerid{redacted}
\fi

\usepackage{subcaption}
\usepackage{placeins}
\usepackage{float}
\usepackage{mathtools}
\usepackage[textsize=tiny]{todonotes}
\definecolor{bagelorange}{HTML}{AE3E06}

\usepackage{arxiv_style}
\setarxivlogo{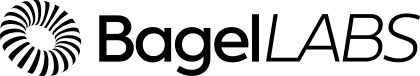}
\setarxivlogowidth{90pt}
\setarxivlogoraise{0pt}
\setarxivdatetext{}

\usepackage[numbers,sort&compress]{natbib}
\hypersetup{
  plainpages=false,
  pdftitle={WorldDiT: A Unified Diffusion Architecture for World and Action Modeling},
  pdfauthor={Sen Wang, R. Gnana Praveen, Bidhan Roy, Marcos Villagra}
}

\theoremstyle{plain}

\theoremstyle{definition}

\theoremstyle{remark}

\title{WorldDiT: A Unified Diffusion Architecture for\\
World and Action Modeling}
\correspondingauthor{research@bagel.com}

\makeatletter
\renewcommand\AB@authnote[1]{}
\renewcommand\AB@affilnote[1]{}
\makeatother
\author{Sen Wang}
\author{R. Gnana Praveen}
\author{Bidhan Roy}
\author{Marcos Villagra}

\newcommand{\worlddithfrepolink}{%
  \raisebox{-0.2ex}[0pt][0pt]{\includegraphics[height=1em]{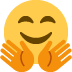}}%
  \,\href{https://huggingface.co/bageldotcom/worlddit}{\textcolor{bagelorange}{\texttt{bageldotcom/worlddit}}}%
}

\affil{Bagel Labs (bagel.com)}

\makeatletter
\renewcommand{\maketitle}{\bgroup\setlength{\parindent}{0pt}
  \begin{adjustwidth}{0pt}{24pt}
    \begin{flushleft}
      {
        {\raggedright \titlefont \@title\par}%
        \vskip11pt
        {\raggedright\Authfont Sen Wang, R. Gnana Praveen, Bidhan Roy and Marcos Villagra\\[\affilsep]
        \Affilfont Bagel Labs (bagel.com)\quad \worlddithfrepolink\par}%
        \vskip20pt%
      }%
    \end{flushleft}
  \end{adjustwidth}
  \egroup
  {%
    {\abscontent}
  }%
  \thispagestyle{firststyle}
}%
\makeatother

\begin{abstract}
Many recent robot policies pursue stronger control by using large pretrained
vision-language models (VLMs) as the action backbone. We introduce WorldDiT, a
unified diffusion transformer architecture that couples action generation with
visual world modeling and achieves strong performance without a large pretrained
VLM action backbone. During training, a single diffusion transformer generates continuous action chunks and
predicts normalized RGB patch targets from future camera frames. Across four
LIBERO simulation suites, WorldDiT lies on the reported Pareto frontier for
total model parameters and mean success among methods reporting all four
suites. These results provide a strong sub-billion-parameter baseline for
future scaling studies.
\end{abstract}

\begin{document}
\fancyhead[C]{\footerfont WorldDiT}
\maketitle

\begin{figure}[H]
\centering
\includegraphics[
  width=\textwidth,
  height=0.34\textheight,
  trim=0 0 0 0,
  clip,
  keepaspectratio
]{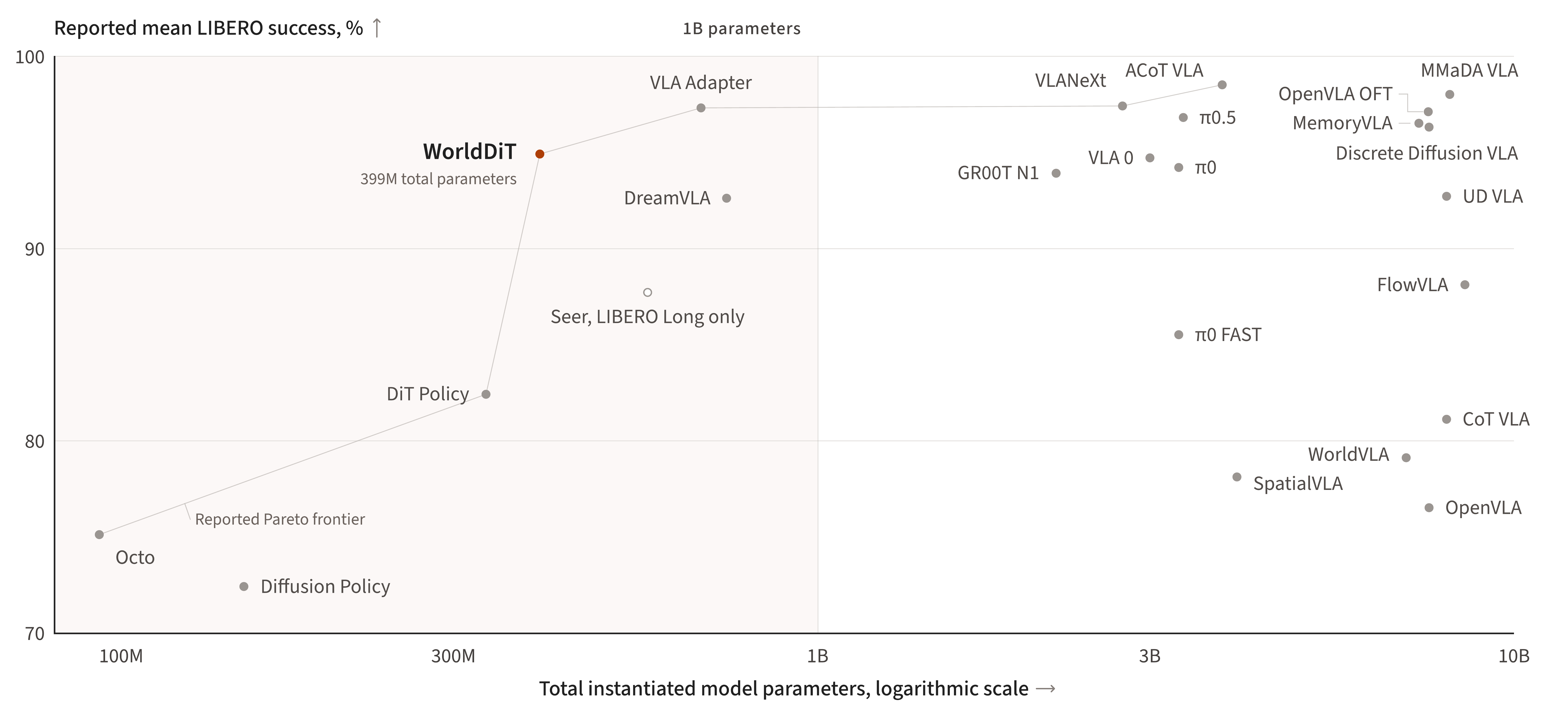}
\caption{Reported LIBERO success against total model parameters for $24$
methods. The line connects methods with mean success computed across all four
suites on the reported Pareto frontier.}
\label{fig:score_vs_param}
\end{figure}

\section{Introduction}

Many robot policies follow a familiar pattern from language modeling, attaching
action generation to a large pretrained model. This pattern brings broad
perception and language understanding, but it also makes the contributions of
scale and architecture difficult to separate. When a policy contains several
billion parameters, strong control may come from the action design, the
pretrained backbone, or their combination. We therefore ask whether a unified
diffusion transformer architecture can combine continuous action generation with an
auxiliary future visual prediction objective and retain strong control
performance without using a large pretrained vision-language model as the
action backbone.

We pursue this objective with WorldDiT, whose frozen visual and language
encoders and trainable robot state encoder condition the shared DiT backbone on
recent visual observations, robot state, and a language instruction. During
training, the shared backbone learns a continuous flow
for a seven-step action chunk and an auxiliary normalized RGB patch target
selected from future primary-camera and wrist-camera frames. At inference, the
shared backbone generates an action chunk. We execute the first three actions, observe
the resulting state, and replan.

\begin{figure}[H]
\centering
\begin{tabular}{@{}c@{\hspace{2pt}}c@{\hspace{2pt}}c@{\hspace{2pt}}c@{\hspace{2pt}}c@{}}
&
{\scriptsize Frame 1} &
{\scriptsize Frame 2} &
{\scriptsize Frame 3} &
{\scriptsize Frame 4} \\
\rotatebox{90}{\scriptsize LIBERO Spatial} &
\includegraphics[width=0.225\textwidth]{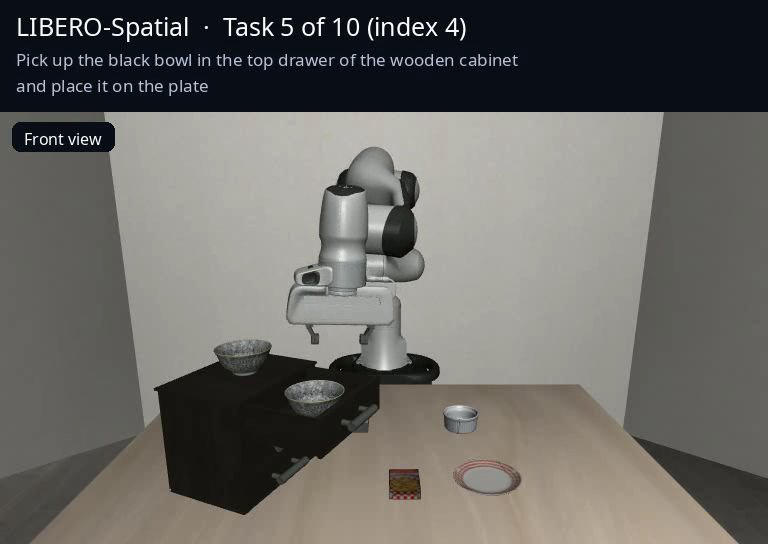} &
\includegraphics[width=0.225\textwidth]{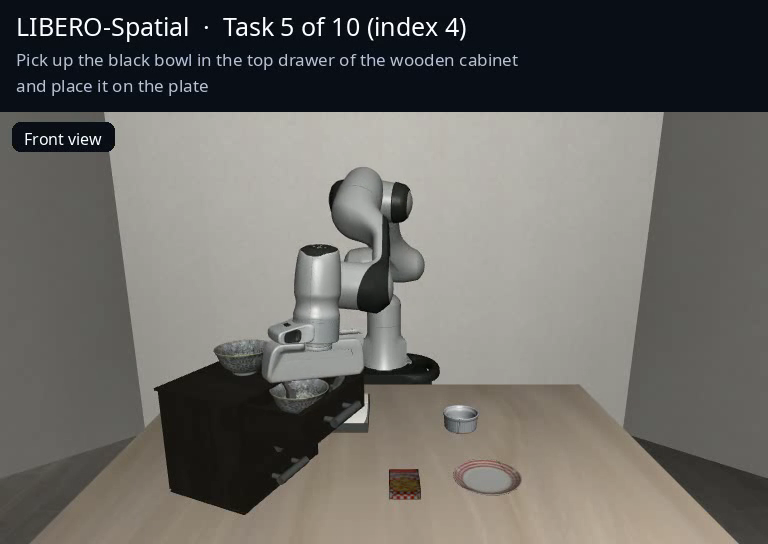} &
\includegraphics[width=0.225\textwidth]{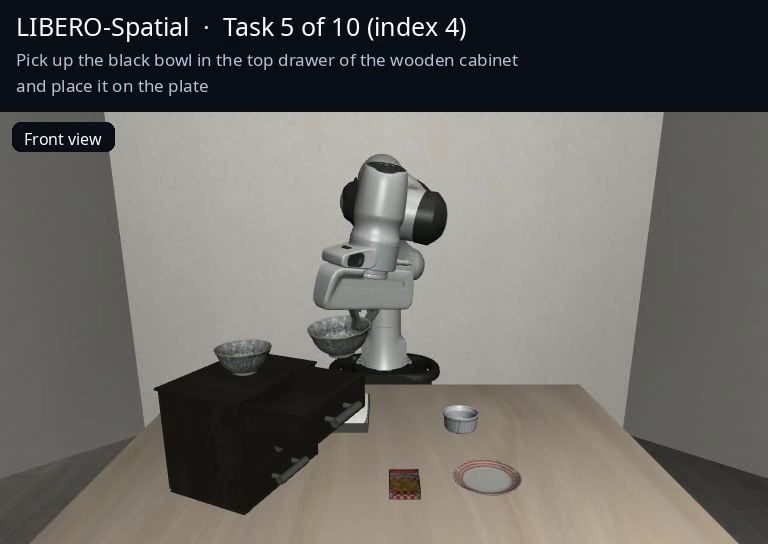} &
\includegraphics[width=0.225\textwidth]{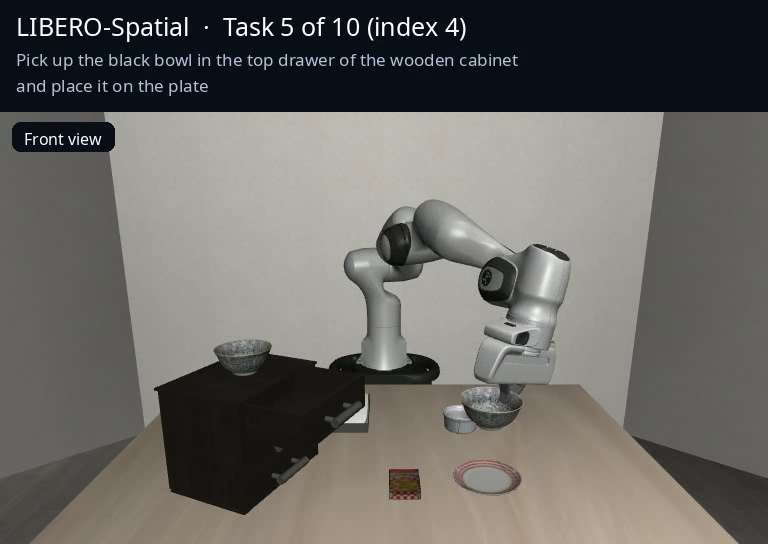} \\
\rotatebox{90}{\scriptsize LIBERO Object} &
\includegraphics[width=0.225\textwidth]{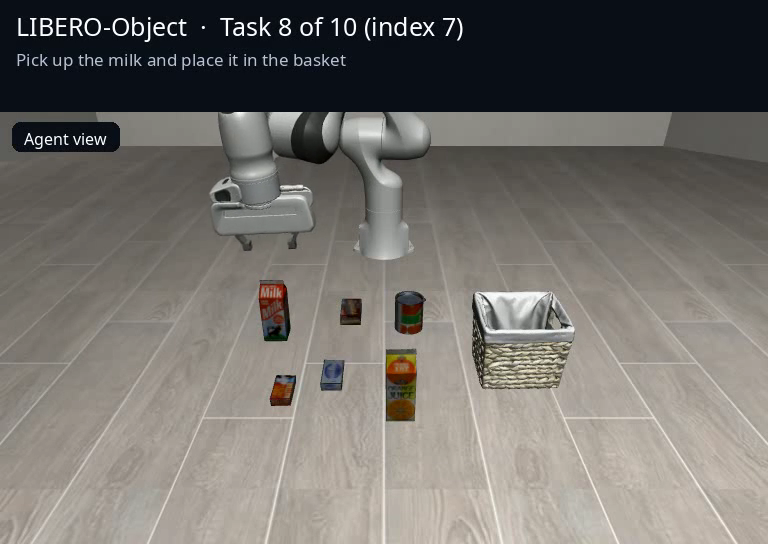} &
\includegraphics[width=0.225\textwidth]{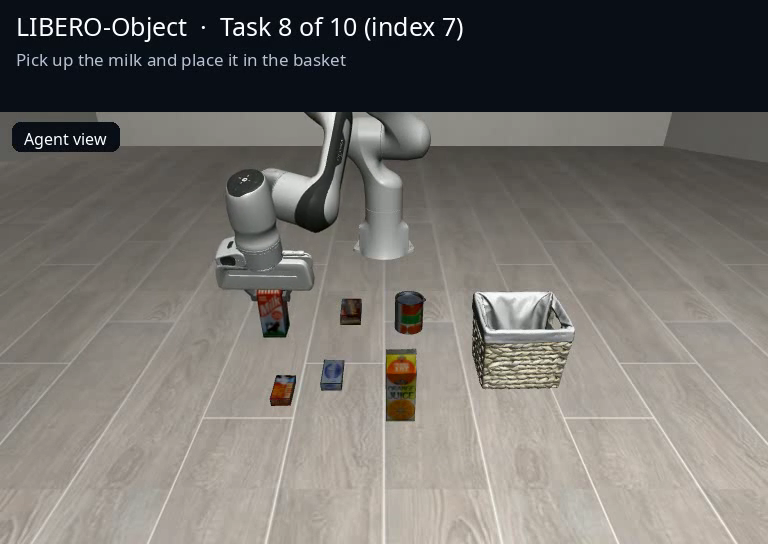} &
\includegraphics[width=0.225\textwidth]{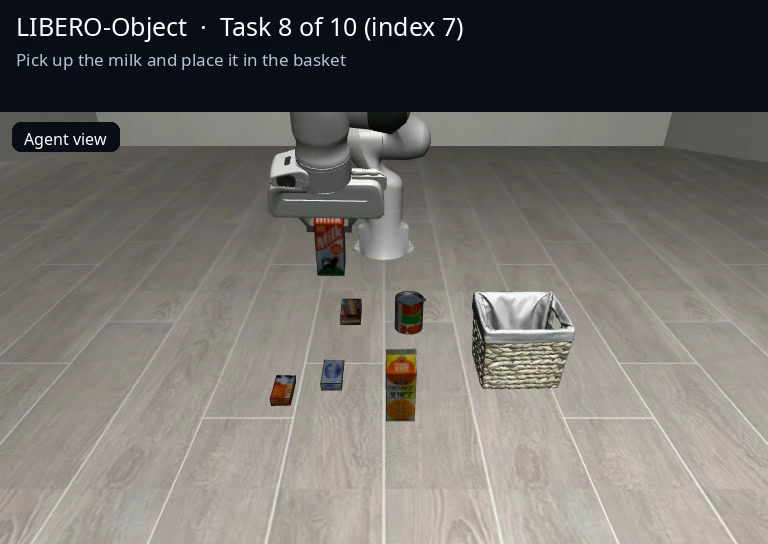} &
\includegraphics[width=0.225\textwidth]{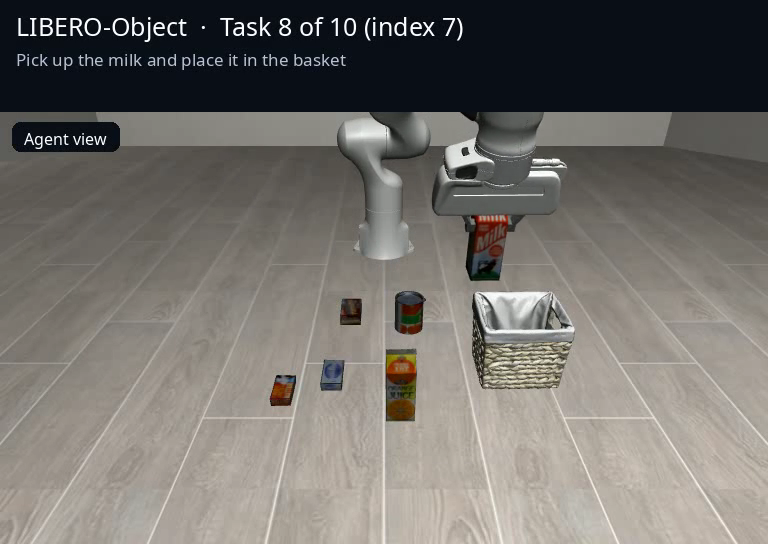} \\
\rotatebox{90}{\scriptsize LIBERO Goal} &
\includegraphics[width=0.225\textwidth]{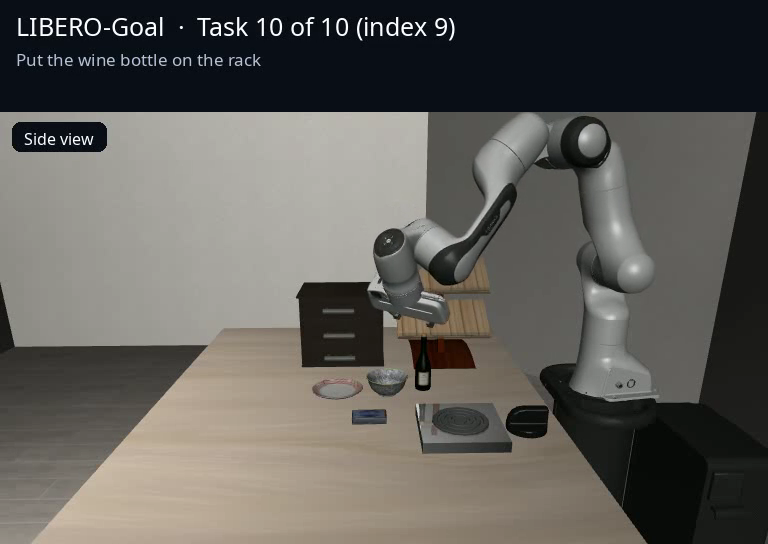} &
\includegraphics[width=0.225\textwidth]{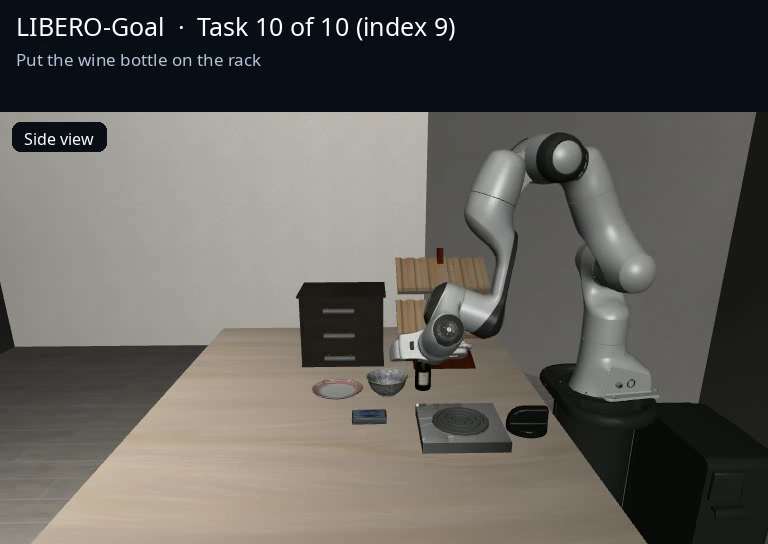} &
\includegraphics[width=0.225\textwidth]{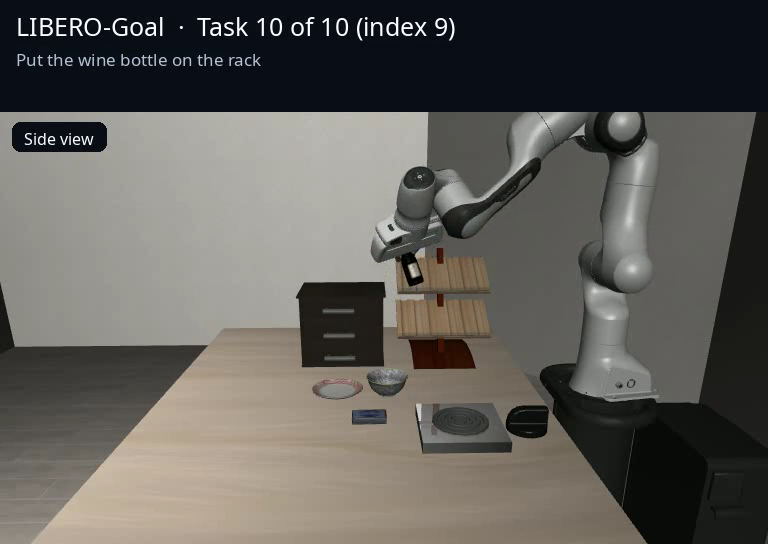} &
\includegraphics[width=0.225\textwidth]{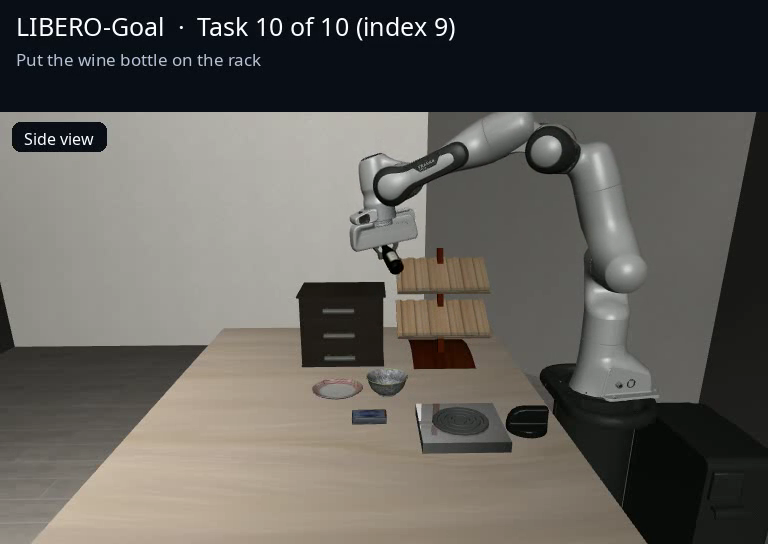} \\
\rotatebox{90}{\scriptsize LIBERO Long} &
\includegraphics[width=0.225\textwidth]{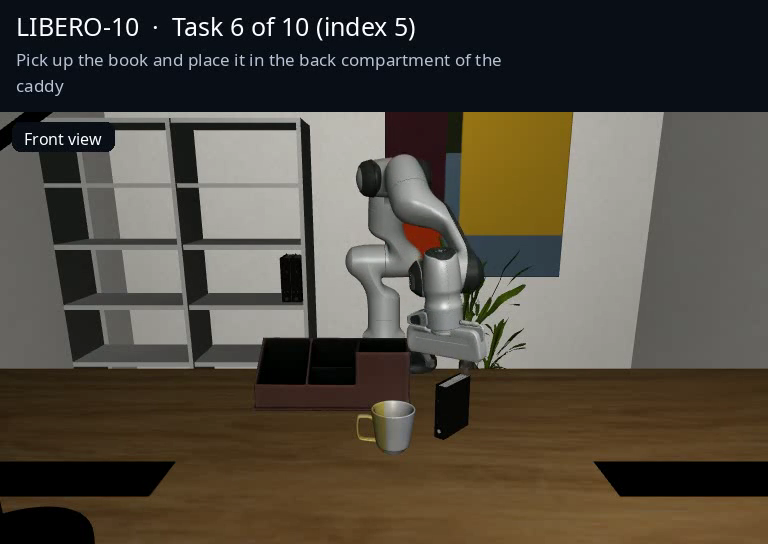} &
\includegraphics[width=0.225\textwidth]{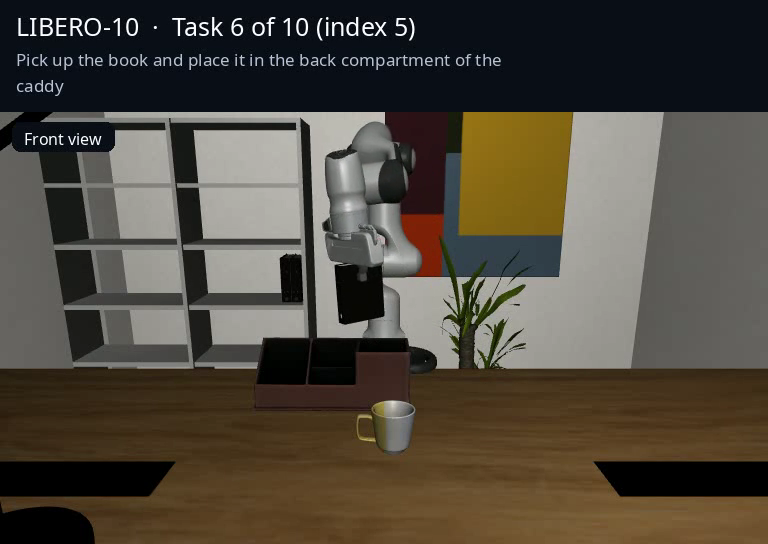} &
\includegraphics[width=0.225\textwidth]{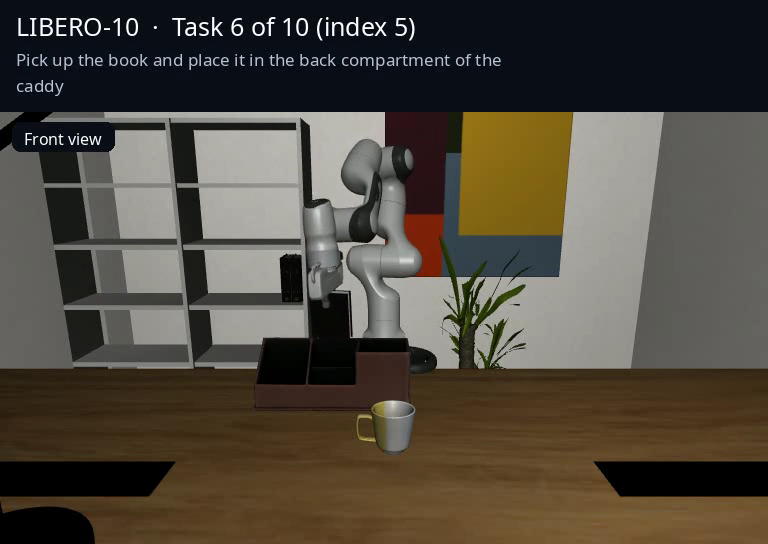} &
\includegraphics[width=0.225\textwidth]{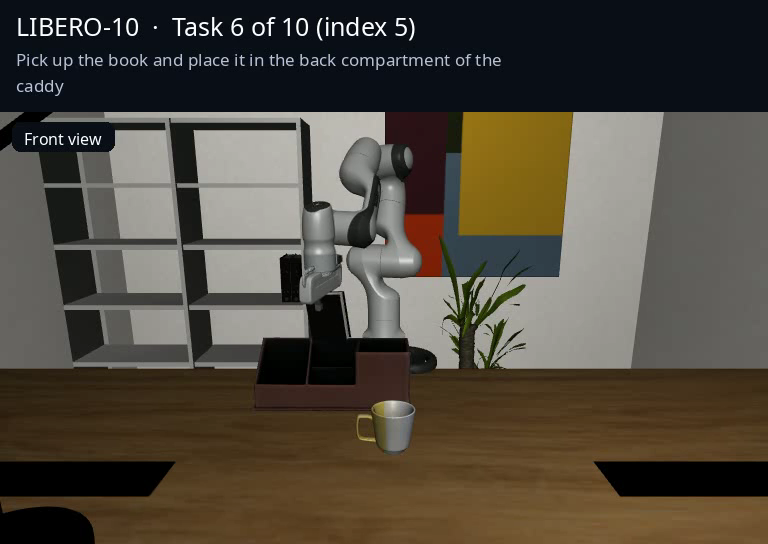}
\end{tabular}
\caption{Each row shows four temporally ordered frames from a successful
WorldDiT rollout in one LIBERO suite.}
\label{fig:qualitative_rollouts}
\end{figure}

Figure~\ref{fig:qualitative_rollouts} shows successful rollouts across four
LIBERO suites.

Figure~\ref{fig:score_vs_param} summarizes the reported parameter and
performance comparison across all $24$ methods. We evaluate one
WorldDiT configuration under the specified training and evaluation protocol,
providing a baseline for future scaling studies rather than evidence of scaling
behavior.

Within this scope, we contribute a unified diffusion transformer architecture
for action generation and future normalized RGB patch prediction, a training and deployment design in
which RGB patch prediction supplies supervision but is absent at inference,
and reported results that characterize the tradeoff between parameter count
and mean success below one billion parameters.

\FloatBarrier
\section{Method}
\label{sec:method}

\subsection{Overview}
We introduce \textbf{WorldDiT}, a unified diffusion transformer architecture that
learns action generation together with an auxiliary future normalized RGB
patch objective. Unlike recent world-action models that rely on a large
vision-language model to generate action tokens autoregressively, WorldDiT uses
one shared DiT backbone to model continuous robot actions and
normalized RGB patches selected from future camera frames.

Given a language instruction and a temporally ordered context of observations
from multiple views and robot states, WorldDiT encodes the observed context
using frozen visual and language encoders. During training, the shared DiT
backbone receives action tokens and normalized RGB patch tokens corrupted with
Gaussian noise, selected from future primary-camera and wrist-camera frames, and
predicts their flow velocities. Future RGB patch prediction supplies auxiliary
supervision during training. Deployment follows the action path directly,
keeping RGB patch tokens and the RGB prediction head outside the inference
graph.

\begin{figure}[!t]
  \centering
  \captionsetup{skip=4pt}
  \includegraphics[width=\textwidth]{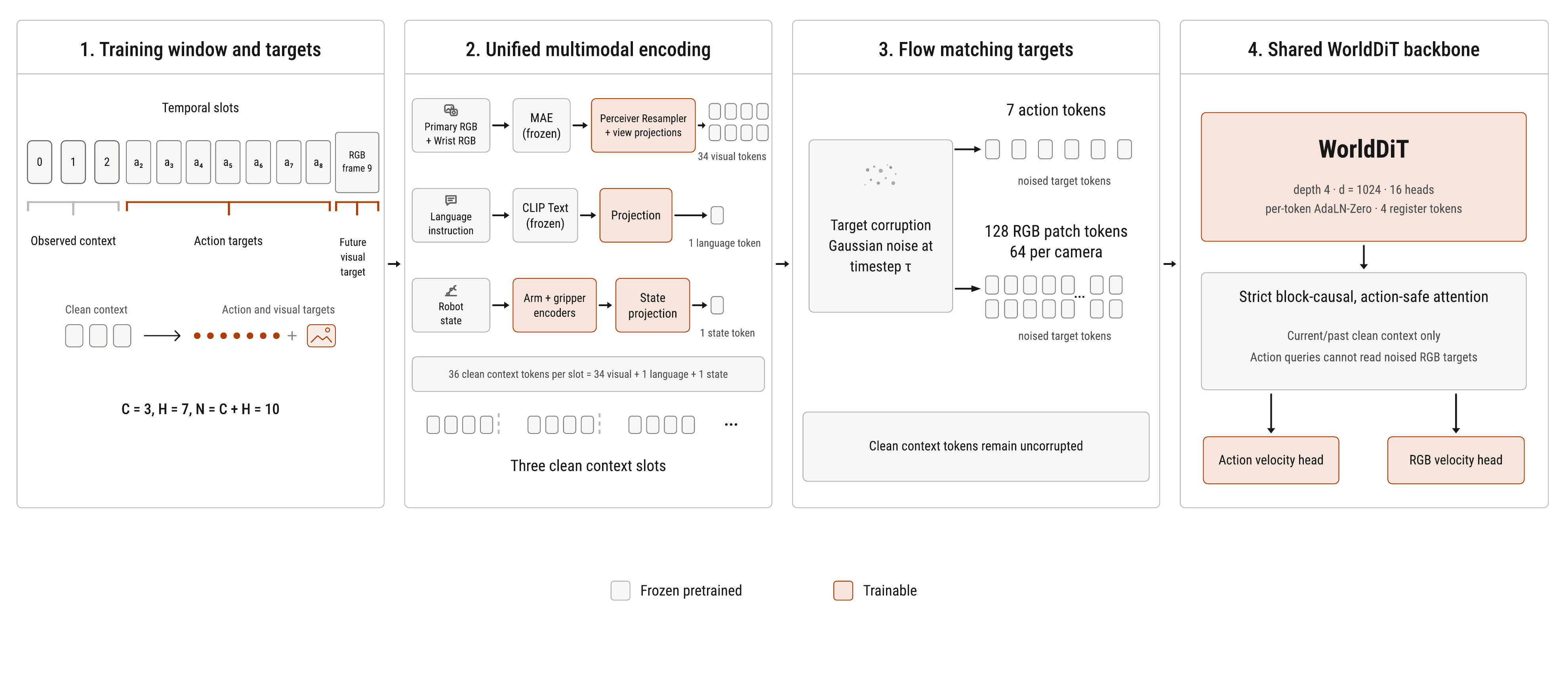}
  \caption{A ten-step window provides three observed context steps, seven target actions, and one future normalized RGB patch target. Frozen encoders and trainable projections construct multimodal tokens, and one unified WorldDiT backbone predicts action and RGB patch flow velocities.}
  \label{fig:training-pipeline}
\end{figure}

Figure~\ref{fig:training-pipeline} summarizes this path from the observation window and multimodal tokenization through the shared backbone to final slot supervision.

Let an $N$-step trajectory segment beginning at index $u$ be
\begin{equation}
\mathcal{W}_{u}
=
\left\{
\left(
\mathbf{o}_{i},
\mathbf{s}_{i},
\mathbf{a}_{i}
\right)
\right\}_{i=u}^{u+N-1},
\end{equation}
where
\begin{equation}
\mathbf{o}_{i}
=
\left(
I_{i}^{\mathrm{p}},
I_{i}^{\mathrm{w}}
\right)
\end{equation}
contains the primary-camera image $I_{i}^{\mathrm{p}}$ and wrist-camera image
$I_{i}^{\mathrm{w}}$, $\mathbf{s}_{i}$ is the robot state, and
$\mathbf{a}_{i}\in\mathbb{R}^{d_a}$ is the robot action. A language instruction
$\ell$ is shared across the trajectory segment.

The first $C$ steps form the observation context. Let
$i=u+C-1$ denote the final observed step and therefore the current control
step. The history available
at position $i$ is
\begin{equation}
\mathcal{H}_{i}
=
\left(
\ell,
\left\{
\left(
\mathbf{o}_{j},
\mathbf{s}_{j}
\right)
\right\}_{j=u}^{i}
\right).
\label{eq:causal-history}
\end{equation}

The corresponding action target is an $H$-step action chunk:
\begin{equation}
\mathbf{A}_{i}
=
\left[
\mathbf{a}_{i},
\mathbf{a}_{i+1},
\ldots,
\mathbf{a}_{i+H-1}
\right]
\in
\mathbb{R}^{H\times d_a}.
\label{eq:action-chunk}
\end{equation}

We construct the future normalized RGB patch target from the primary-camera and
wrist-camera frames at temporal offset $H$. Each CLIP-preprocessed $224$ by $224$
RGB frame is divided into $16$ by $16$ patches. Each $768$-dimensional patch vector
is normalized using its own mean and variance, and $64$ evenly spaced patches
are retained from each camera. Concatenating both camera targets gives
\[
\mathbf{Y}_{i}^{\mathrm{rgb}}
\in
\mathbb{R}^{128\times768}.
\]
The loss is applied only to the final temporal slot, whose RGB patch target
comes from the frame at $i+H$.

Before entering WorldDiT, a trainable linear input projection maps each
corrupted $768$-dimensional RGB patch vector to the backbone hidden dimension,
and a trainable RGB output projection maps each corresponding hidden
representation back to a $768$-dimensional patch velocity. These projections
define the model interface and do not change the supervision target, which
remains the normalized RGB patch vector. For the robot state, we encode the arm
and gripper components separately, concatenate their embeddings, and project
the result into one state token for each temporal slot.

\subsection{Training}
\label{sec:training}
We train the model using flow matching \citep{lipman2023flowmatching}.
The WorldDiT backbone receives the clean context tokens, action tokens and
normalized RGB patch tokens corrupted with Gaussian noise, their timestep
embeddings, and learned register tokens, and predicts the corresponding
velocities.

The training objective regresses the velocity of a straight path from Gaussian
noise to a clean target. For either an action target or a future RGB patch
target,
let $\boldsymbol{\epsilon}\sim\mathcal{N}(\mathbf{0},\mathbf{I})$,
$\mathbf{y}$ denote the clean target, sample
$\tau\sim\mathcal{U}[0,1]$, and define
\[
\mathbf{x}_{\tau}
=
(1-\tau)\boldsymbol{\epsilon}
+
\tau\mathbf{y}.
\]
The flow-matching objective is
\[
\mathcal{L}_{\text{flow}}
=
\mathbb{E}_{\boldsymbol{\epsilon},\mathbf{y},\tau}
\left[
\left\|
v_\theta(\mathbf{x}_{\tau},\tau,c)
-
(\mathbf{y}-\boldsymbol{\epsilon})
\right\|_2^2
\right].
\]
Here $c$ is the context for the current step and
$\mathbf{y}-\boldsymbol{\epsilon}$ is the
target velocity.

WorldDiT predicts one velocity for each target. The total loss is the weighted
sum of the action velocity loss and the RGB patch velocity loss.
\begin{equation}
\mathcal{L}_{\mathrm{total}}
=
w_{\mathrm{action}}
\mathcal{L}_{\mathrm{flow}}^{\mathrm{action}}
+
w_{\mathrm{rgb}}
\mathcal{L}_{\mathrm{flow}}^{\mathrm{rgb}}.
\end{equation}
The coefficients $w_{\mathrm{action}}$ and $w_{\mathrm{rgb}}$ are the
corresponding loss weights.

WorldDiT uses the $C$ observed steps for conditioning, then predicts one
$H$-step action chunk and one future world target composed of normalized RGB
patches from the primary-camera and wrist-camera frames at $i+H$. Only the final
temporal slot contributes to the loss.

\FloatBarrier
\subsection{Inference}
\label{sec:inference}

At deployment, the unified backbone receives a $C$-step history ending at the
current control step $t$:
\begin{equation}
\mathcal{H}_{t}
=
\left(
\ell,
\left\{
\left(
\mathbf{o}_{j},
\mathbf{s}_{j}
\right)
\right\}_{j=t-C+1}^{t}
\right).
\end{equation}

The same multimodal encoding path used during training maps
$\mathcal{H}_t$ to the conditioning representation $\mathbf{G}_t$. No future
RGB patch targets or future action labels are provided.

\begin{figure}[!t]
  \centering
  \includegraphics[width=\textwidth]{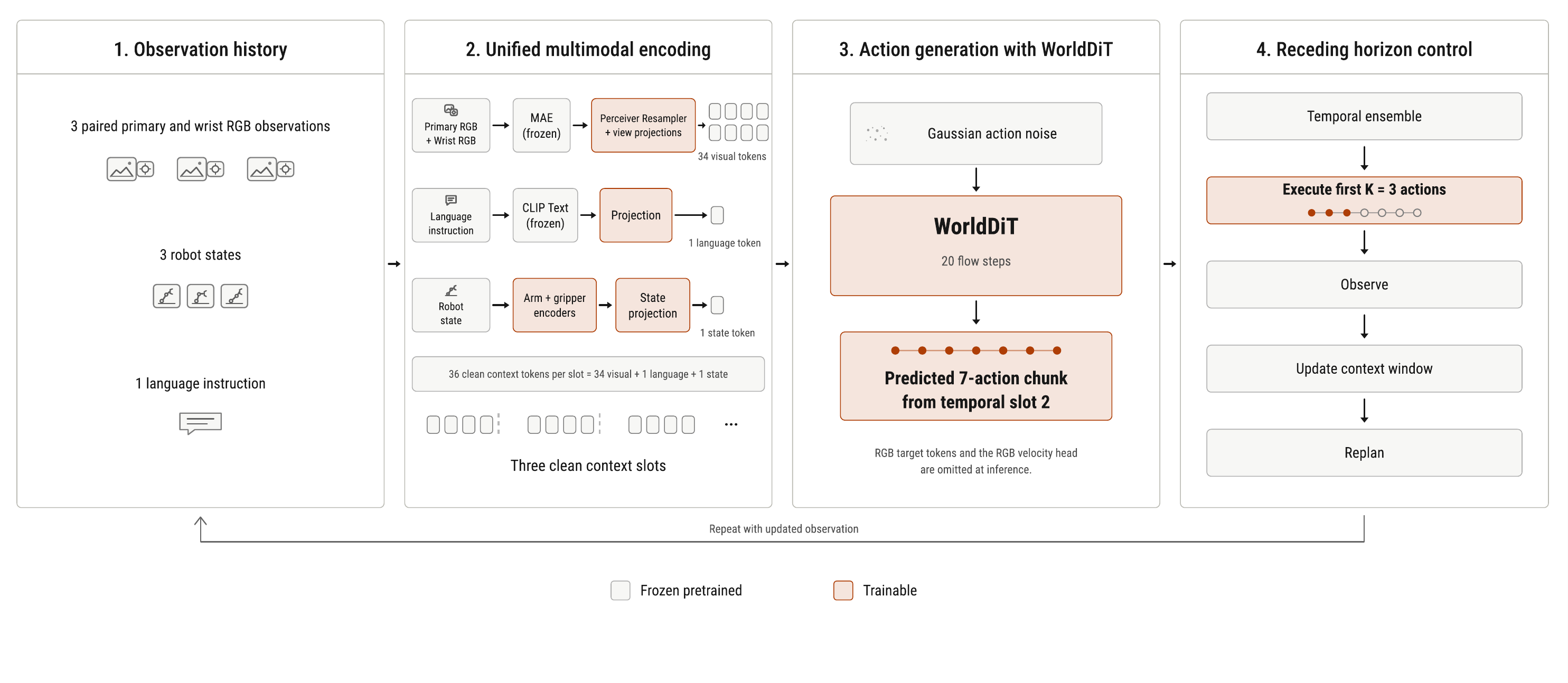}
  \caption{The unified backbone encodes three observed context steps and samples a chunk of seven actions through twenty flow steps. We execute the first three actions, update the window, and replan.}
  \label{fig:inference-pipeline}
\end{figure}

Figure~\ref{fig:inference-pipeline} summarizes this deployment path from observed history through action generation and receding-horizon control.

Action generation begins from Gaussian noise:
\begin{equation}
\mathbf{x}_{t,0}^{\mathrm{act}}
\sim
\mathcal{N}
\left(
\mathbf{0},
\mathbf{I}
\right).
\end{equation}

The learned action velocity field is then numerically integrated:
\begin{equation}
\frac{
d\mathbf{x}_{t,\tau}^{\mathrm{act}}
}{
d\tau
}
=
v_{\theta}^{\mathrm{act}}
\left(
\mathbf{x}_{t,\tau}^{\mathrm{act}},
\tau,
\mathbf{G}_{t}
\right),
\qquad
\tau\in[0,1].
\label{eq:inference-flow}
\end{equation}

After $T_{\mathrm{samp}}$ integration steps, the final flow state defines the
predicted action chunk:
\begin{equation}
\widehat{\mathbf{A}}_{t}
=
\mathbf{x}_{t,1}^{\mathrm{act}}
\in
\mathbb{R}^{H\times d_a}.
\end{equation}

We execute a prefix of $\widehat{\mathbf{A}}_t$, update the history, and replan.

\section{Experiments}

\subsection{Setup}

\subsubsection{Data and model configuration}
We evaluate WorldDiT on the LIBERO manipulation benchmark~\citep{liu2023libero}. We use the
four downstream suites \texttt{libero\_spatial}, \texttt{libero\_object},
\texttt{libero\_goal}, and \texttt{libero\_10}, reported as LIBERO Long, together
with the large multi-task \texttt{libero\_90} split used only for pretraining.
We convert raw demonstrations to fixed-length windows containing multi-view
RGB observations, robot states, language instructions, and action sequences.
Future primary-camera and wrist-camera frames in each window provide the auxiliary
normalized RGB patch targets.

Visual observations, language instructions, and robot states are encoded by a
frozen MAE image encoder, a shared Perceiver Resampler, a frozen CLIP text
encoder, and a trainable robot state encoder. The resulting tokens are
concatenated and used to condition WorldDiT.

The WorldDiT backbone has depth $4$, hidden size $1024$, $16$ attention heads,
and $4$ register tokens. Each input to the unified backbone contains a
temporally ordered context of $C=3$ observations. The unified backbone predicts
an $H=7$ action chunk,
$\mathbf{A}_i\in\mathbb{R}^{7\times 7}$, and uses $64$ normalized RGB patch
tokens from each future camera frame, for $128$ target tokens in total. The
complete training window contains $N=10$ steps. We train this configuration
solely with the unified flow-matching objective of
Section~\ref{sec:training}.

\subsubsection{Training}
On LIBERO we first pretrain on the multi-task \texttt{libero\_90} split for
$30$ epochs with a per-GPU batch size of $40$ and gradient accumulation of $2$,
a learning rate of $1\times10^{-4}$ with a cosine schedule and one warmup epoch,
sequence length $10$, observation window $10$, a context length of $3$ observations and a complete data-window length of $10$ steps.

For LIBERO we fine-tune the pretrained \texttt{libero\_90} checkpoint
\emph{independently on each downstream suite}. Fine-tuning uses an effective
batch size of $512$, with a per-GPU batch size of $32$ and two gradient
accumulation steps on eight GPUs. The unified head is trained on
action and normalized RGB patch targets, with action and RGB patch loss weights
of $0.1$ and $0.001$ respectively, with a learning rate of $1\times10^{-4}$ in
bf16 mixed precision.

We conduct all training runs on one node with eight RTX Pro 6000 GPUs.

\subsubsection{Evaluation}
Under this protocol, we evaluate one WorldDiT configuration, fine-tuned
separately for each of four suites, over $500$ episodes per suite.

At inference there is no clean target. We initialize the action tokens from
Gaussian noise and integrate the predicted velocity field from $\tau{=}0$ to
$\tau{=}1$ with $20$ Euler steps,
$x_{\tau+\Delta\tau}=x_\tau+\Delta\tau\,v_\theta(x_\tau,\tau,c)$, decoding the
resulting action
chunk and sliding the context window. All simulator rollouts use headless
EGL rendering. On LIBERO we evaluate each suite with action ensembling and
report success rates for each task and their mean over the suite's tasks. The unified
backbone predicts seven actions. We execute three actions before replanning and
apply temporal ensembling to overlapping absolute-time action predictions.

For each suite, the reported WorldDiT score aggregates $500$ simulator
episodes. The aggregate includes $300$ episodes per suite used during staged
checkpoint selection. It is therefore not fully held out, and the reported
$94.9\%$ should not be interpreted as an unbiased test estimate. The incomplete
public availability of model weights and training code prevents us from
reproducing every compared method under a shared protocol. We therefore use
scores from the cited reports. The graph and table use total instantiated parameter counts,
including frozen modules required at inference, and label reconstructed counts
as approximate.

\FloatBarrier
\subsection{Results}

We evaluate a separately fine-tuned WorldDiT model on each of the four LIBERO
suites over $500$ simulator episodes.
WorldDiT records
$98.0\%$ on Spatial, $97.0\%$ on Object, $92.8\%$ on Goal, and $91.8\%$ on
Long, producing a mean success rate of $94.9\%$. Long remains the hardest
suite, which is consistent with the extended multistage behavior required by
its tasks.

Figure~\ref{fig:score_vs_param} plots the same
$24$ rows against total instantiated parameter count. Seer is shown at its
reported Long score but is excluded from the Pareto calculation because a
mean over all four suites is unavailable.

\vspace{0.5\baselineskip}
\begin{table}[H]
    \centering
    \captionsetup{labelformat=empty,labelsep=none}
    \caption{Table~\ref{tab:performance_comparison} shows the reported
    performance of all $24$ methods across the LIBERO Spatial, Object, Goal,
    and Long suites. Methods are grouped by whether they use a large pretrained
    vision-language model as the action backbone. The WorldDiT row reports an
    aggregate over $500$ episodes per suite, including $300$ episodes per suite
    used during staged checkpoint selection.}
    \label{tab:performance_comparison}
    \small
    \setlength{\tabcolsep}{3pt}
    \begin{tabular}{llccccc}
        \toprule
        \textbf{Category} & \textbf{Method} & \textbf{Spatial} & \textbf{Object} & \textbf{Goal} & \textbf{Long} & \textbf{Mean} \\
        \midrule
        \raisebox{-5\baselineskip}[0pt][0pt]{Large pretrained VLM action backbone} & ACoT VLA~\citep{zhong2026acot} & 98.6 & 99.0 & 99.4 & 97.0 & 98.5 \\
         & MMaDA VLA~\citep{liu2026mmada} & 98.8 & 99.8 & 98.0 & 95.2 & 98.0 \\
         & VLANeXt~\citep{wu2026vlanext} & 99.0 & 99.2 & 96.6 & 94.8 & 97.4 \\
         & VLA Adapter~\citep{wang2026vlaadapter} & 97.8 & 99.2 & 97.2 & 95.0 & 97.3 \\
         & OpenVLA OFT~\citep{kim2025oft} & 97.6 & 98.4 & 97.9 & 94.5 & 97.1 \\
         & $\pi_{0.5}$~\citep{physicalintelligence2025pi05} & 98.8 & 98.2 & 98.0 & 92.4 & 96.9 \\
         & MemoryVLA~\citep{shi2026memoryvla} & 98.4 & 98.4 & 96.4 & 93.4 & 96.7 \\
         & Discrete Diffusion VLA~\citep{liang2025ddvla} & 97.2 & 98.6 & 97.4 & 92.0 & 96.3 \\
         & VLA 0~\citep{goyal2025vla0} & 97.0 & 97.8 & 96.2 & 87.6 & 94.7 \\
         & $\pi_0$~\citep{black2024pi0} & 96.8 & 98.8 & 95.8 & 85.2 & 94.2 \\
         & GR00T N1~\citep{bjorck2025groot} & 94.4 & 97.6 & 93.0 & 90.6 & 93.9 \\
         & UD VLA~\citep{chen2026udvla} & 94.1 & 95.7 & 91.2 & 89.6 & 92.7 \\
         & $\pi_0$ FAST~\citep{PertschK-RSS-25} & 96.4 & 96.8 & 88.6 & 60.2 & 85.5 \\
         & CoT VLA~\citep{zhao2025cotvla} & 87.5 & 91.6 & 87.6 & 69.0 & 83.9 \\
         & WorldVLA~\citep{cen2025worldvla} & 85.6 & 89.0 & 82.6 & 59.0 & 79.1 \\
         & SpatialVLA~\citep{qu2025spatialvla} & 88.2 & 89.9 & 78.6 & 55.5 & 78.1 \\
         & OpenVLA~\citep{kim2025openvla} & 84.7 & 88.4 & 79.2 & 53.7 & 76.5 \\
        \midrule
         \raisebox{-5.25\baselineskip}[0pt][0pt]{No large pretrained VLM action backbone} & \textbf{WorldDiT} & \textbf{98.0} & \textbf{97.0} & \textbf{92.8} & \textbf{91.8} & \textbf{94.9} \\
         & DreamVLA~\citep{dreamvla} & 97.5 & 94.0 & 89.5 & 89.5 & 92.6 \\
         & FlowVLA~\citep{zhong2025flowvla} & 93.2 & 95.0 & 91.6 & 72.6 & 88.1 \\
         & DiT Policy~\citep{hou2024ditpolicy} & 84.2 & 96.3 & 85.4 & 63.8 & 82.4 \\
         & Octo~\citep{octo2024} & 78.9 & 85.7 & 84.6 & 51.1 & 75.1 \\
         & Diffusion Policy~\citep{chi2023diffusionpolicy} & 78.3 & 92.5 & 68.3 & 50.5 & 72.4 \\
         & Seer~\citep{seer} & N/A  & N/A  & N/A  & 87.7 & N/A \\
        \bottomrule
    \end{tabular}
\end{table}

WorldDiT uses $399.084$ million total parameters, including frozen modules
required at inference, and $135.107$ million trainable parameters. This point
lies on the Pareto frontier in the reported comparison. Among methods with
a reported mean over all four suites, every method with a higher reported success value
uses more total parameters, while every other method at or below the WorldDiT
parameter count reports lower success. This characterizes the tradeoff between
parameter count and reported mean success among the published results included
in the comparison.

\FloatBarrier
\section{Discussion}
\label{sec:discussion}

WorldDiT shows that a single diffusion transformer can couple continuous action
generation with future visual prediction while retaining an action-only
deployment path. Its reported LIBERO performance places the 399M-parameter
system on the parameter--success Pareto frontier, indicating that strong
benchmark performance is attainable without placing a large pretrained
vision-language model in the action backbone. More broadly, this result supports
unified world-and-action modeling as a compact basis for scaling across model
capacity, data diversity, and deployment settings. Future work can characterize
how the normalized future RGB objective shapes control and whether the same
tradeoff persists at larger scales. Since action and visual targets share a
flow-matching formulation, WorldDiT may also support decomposition into
independently trained experts, following the direction explored in Paris and
Paris 2.0, including configurations suited to heterogeneous hardware and
compute-constrained robot platforms~\citep{jiang2025paris,rouzbayani2026paris2}.

\bibliography{references}
\bibliographystyle{icml2026}

\end{document}